\documentclass[letterpaper, 10 pt, journal, twoside]{IEEEtran}

\usepackage{graphics} 
\usepackage{epsfig} 
\usepackage{times} 
\usepackage{amsmath} 
\usepackage{amssymb}  
\usepackage{cite}

\usepackage{multirow}
\usepackage{makecell}
\usepackage{url}
\usepackage{booktabs}
\usepackage{tabularx}
\usepackage{xcolor}
\usepackage{orcidlink}

\begin{document}

\title{\LARGE \bf
LeCropFollow: Latent Space Planning for Navigation\\
in Unstructured Crop Fields
}

\author{Felipe Tommaselli$^{1}$~\orcidlink{0000-0002-9638-4306}, 
        Francisco Affonso$^{2}$~ \orcidlink{0000-0002-8888-1089}, 
        Arthur Pompeu$^{1}$~ \orcidlink{0009-0006-9649-2047}, 
        Gianluca Capezzuto$^{1}$~ \orcidlink{0000-0002-5796-9846},\\
        Arun Narenthiran Sivakumar$^{2}$~ \orcidlink{0000-0001-8711-9431}, 
        Girish Chowdhary$^{2}$~ \orcidlink{0000-0002-4657-307X}, 
        and Marcelo Becker$^{1}$~ \orcidlink{0000-0002-7508-5817}%
        
\thanks{Manuscript received: February 18, 2026; Revised May 21, 2026; Accepted June 11, 2026.}%
\thanks{This paper was recommended for publication by Editor Soon-Jo Chung upon evaluation of the Associate Editor and Reviewers' comments. This work was supported in part by the S\~ao Paulo Research Foundation (FAPESP), Grants \#2022/08330-9, \#2023/15926-8, \#2023/17678-1, \#2024/09442-0, \#2025/20858-7 and \#2025/22381-3; and in part by the Conselho Nacional de Desenvolvimento Cient\'ifico e Tecnol\'ogico (CNPq), Grant 308092/2020-1.}
\thanks{$^{1}$Felipe Andrade G. Tommaselli, Arthur Pompeu, Gianluca Capezzuto, and Marcelo Becker are with the Mobile Robotics Group, Center for Robotics (CRob), S\~ao Carlos School of Engineering (EESC), University of S\~ao Paulo, S\~ao Carlos, SP, Brazil.
}
\thanks{$^{2}$Francisco Affonso, Arun Narenthiran Sivakumar, and Girish Chowdhary are with the DASLab, Department of Computer Science, University of Illinois Urbana-Champaign, Champaign, IL, USA.
}%
\thanks{ Corresponding author: Felipe Tommaselli ({\tt\footnotesize f.tommaselli@usp.br}).
}
\thanks{This work is licensed under a Creative Commons Attribution 4.0 License (CC BY 4.0).}
}

\markboth{IEEE Robotics and Automation Letters. Preprint Version. Accepted June, 2026}
{Tommaselli \MakeLowercase{\textit{et al.}}: LeCropFollow: Latent Space Planning for Navigation in Unstructured Crop Fields} 

\maketitle

\begin{abstract}

Unstructured navigational features, such as irregular planting or discontinuities, remain the primary failure mode for under-canopy agricultural robots. Existing geometric approaches often fail in these scenarios because they compress high-dimensional visual data into deterministic spatial references, effectively discarding the uncertainty and semantic context required to navigate ambiguous terrain. To address this, we present LeCropFollow, a visual navigation framework that bypasses explicit geometric modeling in favor of a learned latent representation. By integrating a self-supervised semantic heatmap extractor with TD-MPC2, a Model-Based Reinforcement Learning (MBRL) planner, our system optimizes trajectories directly within a latent manifold. The framework operates over the uncompressed heatmap signal, preserving the semantic context that geometric reductions discard. We demonstrate that this representational shift enables zero-shot transfer from simplified simulation to the physical world without fine-tuning. Extensive field experiments in late-stage corn fields show that LeCropFollow matches state-of-the-art baselines in unstructured rows but significantly outperforms them in plantation gaps, achieving a 2.4$\times$ reduction in semantic failures compared to keypoint-based methods. These results suggest that latent planning offers a robust alternative to geometric estimation for operations in heterogeneous agricultural environments. Code, models, and data available: \url{https://felipe-tommaselli.github.io/lecropfollow/}.

\end{abstract}

\begin{IEEEkeywords}
Robotics and Automation in Agriculture and Forestry, Representation Learning, Field Robots
\end{IEEEkeywords}

\IEEEpeerreviewmaketitle

\section{INTRODUCTION}

\IEEEPARstart{A}{utonomous} robotic systems are essential for scaling plant-level precision tasks such as phenotyping and mechanical weeding \cite{debruin2025breaking}. To execute these tasks, compact robots must navigate the narrow, GPS-denied spaces beneath the crop canopy \cite{araus2022crop}. In these occluded environments, standard GNSS-RTK solutions degrade due to signal multipath errors \cite{velasquez2022multisensor}. 

To overcome this, the research community widely converged on onboard perception using LiDAR or cameras to estimate local position without reliance on external satellite signals. Whether processing 3D point clouds from LiDAR \cite{affonso2025crow, pinto2023navigating} or RGB streams from cameras \cite{sivakumar2024cropfollowpp}, the standard approach extracts explicit geometric cues, such as crop lines or vanishing points, to construct a reference path. This path subsequently serves as the reference for optimal constrained control in real-time.

Despite significant progress in general under-canopy autonomy, these geometric assumptions are often violated. Agricultural fields are inherently unstructured environments characterized by irregular planting, erosion, and significant vegetation gaps \cite{sivakumar2021learned}. As reported in \cite{sivakumar2024cropfollowpp}, approximately half of autonomy failures stem from unstructured features, such as plantation gaps where row references vanish, and the geometric projection becomes ill-posed. In these scenarios, the controller receives noisy references, leading to divergent behaviors and collisions.

We argue that these failures fundamentally arise from information over-compression. Although modern learning-based perception backbones extract rich environmental features, these are reduced to low-dimensional spatial references for downstream controllers, discarding critical information. Such representations cannot distinguish a clear path from a sensing failure, nor can they capture the semantics of a traversable gap. To achieve robust operation in unstructured environments, navigation policies must retain access to both the uncertainty and semantic context embedded in the observations.

\begin{figure}[t]
    \centering
    \includegraphics[width=\linewidth]{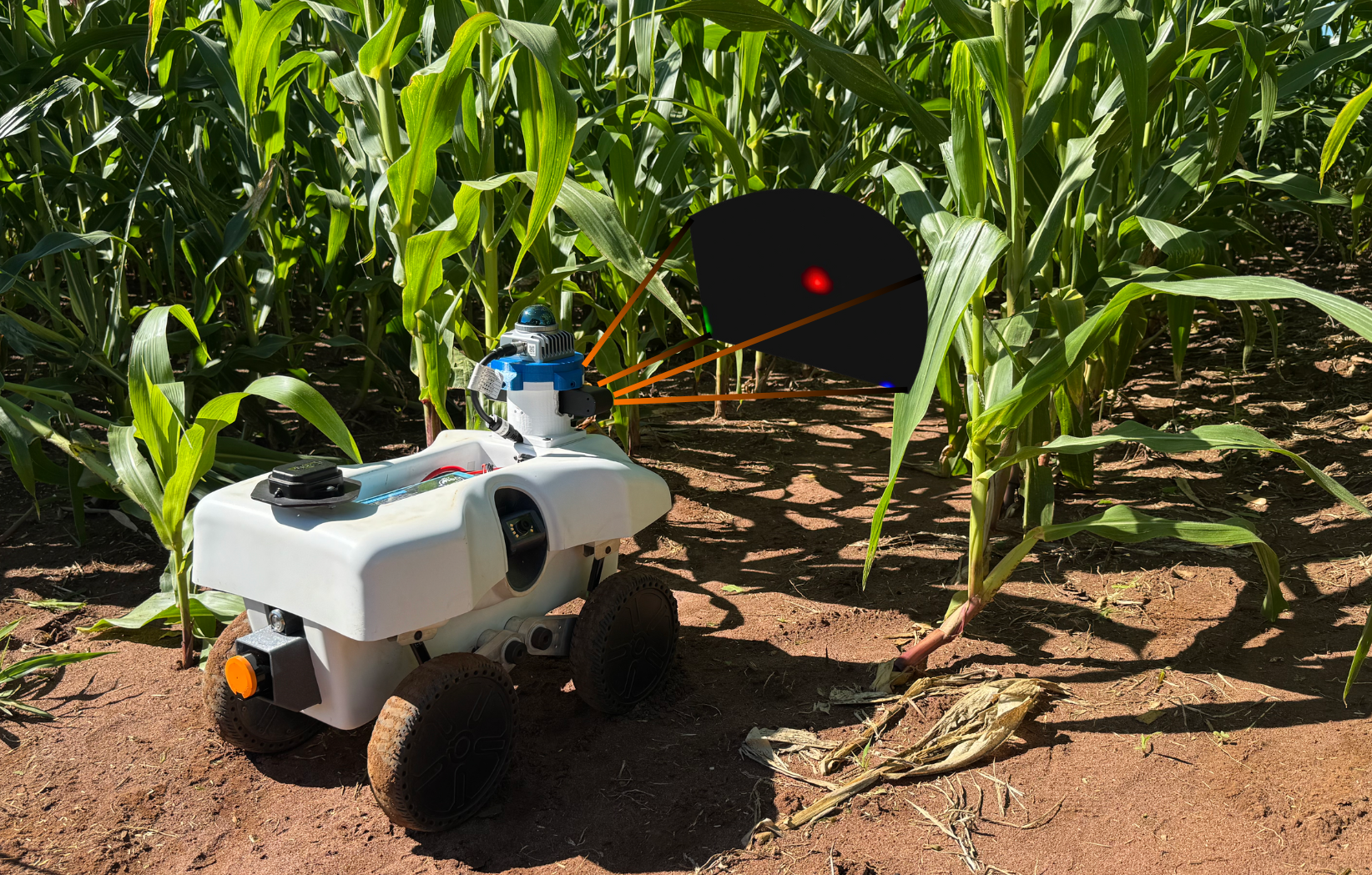}
\caption{\textbf{LeCropFollow.} A learning-based navigation framework for under-canopy agricultural robots that plans trajectories within a learned latent world model over the uncompressed heatmap signal, enabling zero-shot navigation of unstructured fields without GNSS.}
    \label{fig:cover}
\end{figure}

Reinforcement learning (RL) provides a principled alternative by mapping high-dimensional observations directly to actions, allowing the policy to retain uncertainty and semantic context~\cite{sutton1991dyna}. However, model-free RL typically requires extensive on-policy data to cover the full state space, and its policy updates depend on return signals that are non-stationary and highly variable, making learning unreliable for rare but critical events such as crop gaps. A learned world model addresses both limitations. Unlike policy objectives, the world model is trained with a next-state prediction loss that does not depend on reward estimation, providing a more stable learning signal from imbalanced data. At deployment, the model evaluates candidate actions through imagined rollouts in a compact latent space, refining the policy output in scenarios where the learned policy alone would fail~\cite{hafner2023dreamerv3}.

In this work, we introduce \textbf{LeCropFollow} (\textbf{L}atent \textbf{e}mbedding \textbf{CropFollow}), a visual navigation framework that bypasses explicit geometric state estimation, illustrated in Fig. \ref{fig:cover}. We leverage TD-MPC2 \cite{hansen2023tdmpc2}, a Model-Based Reinforcement Learning (MBRL) algorithm, to learn a latent world model. This enables the robot to perform online planning entirely within a learned latent manifold. By optimizing trajectories based on a learned value function, our system maintains stability in heterogeneous crop settings. We demonstrate that this representational shift enables zero-shot transfer from simplified simulation to real-world deployment. Our specific contributions are:

\begin{itemize}
    \item A latent-planning navigation framework that avoids the information over-compression of geometric perception pipelines by keeping the full heatmap signal inside learned components;
    \item Evidence that retaining heatmap dispersion via the latent encoder consistently reduces failures in unstructured crop gaps compared to geometric baselines;
    \item A sim-to-real recipe that enables zero-shot deployment on real-world crops.
\end{itemize}

\section{RELATED WORK}

\textbf{Perception for Under-Canopy Navigation} traditionally relies on geometric distance-based approaches (LiDAR \cite{affonso2025crow, pinto2023navigating, velasquez2022multisensor} or ultrasonic arrays \cite{corno2021adaptive}) for row following. While effective in uniform fields, these modalities struggle to scale on heterogeneous plantations without the semantic information of RGB images \cite{sivakumar2021learned}. To address semantic ambiguity, the field shifted toward vision-based learning approaches, most notably CropFollow \cite{sivakumar2021learned} and CropFollow++ \cite{sivakumar2024cropfollowpp}. Inspired by \cite{vecerik2021s3k}, these methods utilize self-supervised learning to predict plantation keypoints. However, this explicit reliance on keypoint extraction creates a bottleneck in unstructured environments. In scenarios such as plantation gaps, where visual features become ambiguous, the forced prediction of deterministic keypoints yields noisy references that mislead the controller, remaining the primary source of navigational failure \cite{sivakumar2024cropfollowpp}.

\textbf{Heatmap Representations} are frequently utilized within the semantic keypoint paradigm, which explicitly extends classical concepts from visual robotic manipulation \cite{florence2018denseobjectnets, manuelli2019kpam, qin2019keto} where sparse pixel coordinates are extracted to parameterize grasping candidates. In these frameworks, the trained model outputs a spatial probability heatmap, and the specific keypoint coordinate is extracted as the mode of this distribution. While the heatmap inherently encodes uncertainty as Gaussian variance, this extraction step still compresses the full probabilistic signal into a single deterministic coordinate. We posit that this specific information loss is the primary failure mode in unstructured fields; without access to the uncertainty encoded in the heatmap's dispersion, the geometric controller acts with unwarranted confidence in ill-posed scenarios.

\textbf{Visual Representation Learning} remains a long-standing objective in the broader robotics domain, particularly for visuomotor policies that operate directly on pixels \cite{levine2016visumotor}. While raw RGB images inherently retain the rich semantic information necessary for generalizable behavior, learning directly in this representation space is notoriously challenging \cite{nair2023r3m}. These challenges are amplified with the pursuit of world models for planning, where generative approaches attempt to reconstruct full-frame pixel dynamics to predict future outcomes \cite{zhang2025videomodels}. In this work, we show that heatmaps offer a computationally efficient alternative. By predicting feature maps rather than raw pixels, our world model achieves real-time performance while retaining semantic information for control.

\textbf{Model-Based Reinforcement Learning (MBRL)} refers to formulations that leverage either known or learned world models to improve training efficiency through synthetic data generation~\cite{affonso2025learningwalklessdynastyle} or to enhance decision-making via planning~\cite{hafner2023dreamerv3, khurana2025dinowm}. While recent work, such as \cite{mineiro2025endtoend}, demonstrates that model-free approaches are viable for under-canopy navigation, these methods do not explicitly exploit the advantages of MBRL-based planning to improve decision-making over short and bounded horizons. In this work, we build upon the TD-MPC2 framework~\cite{hansen2023tdmpc2}, which enables planning directly in a learned latent space. This design choice complements our heatmap abstraction and allows us to align with and adapt recent advances in efficient representation learning~\cite{zhang2025atk, spahn2025geometric, yin2025rapidlyadapting, wen2024anypoint}.

\section{METHODS}

We propose \textbf{LeCropFollow}, a navigation framework that maps high-dimensional visual observations directly to control inputs without explicit state estimation. At inference time, the perception backbone (trained via self-supervised learning) and the control policy (trained via model-based reinforcement learning) are deployed zero-shot in the physical field. Unlike classical approaches, we do not assume explicit geometric priors; instead, our system performs trajectory planning in a learned latent space with a trained world model.

\begin{figure*}[t]
    \centering
    \includegraphics[width=\linewidth]{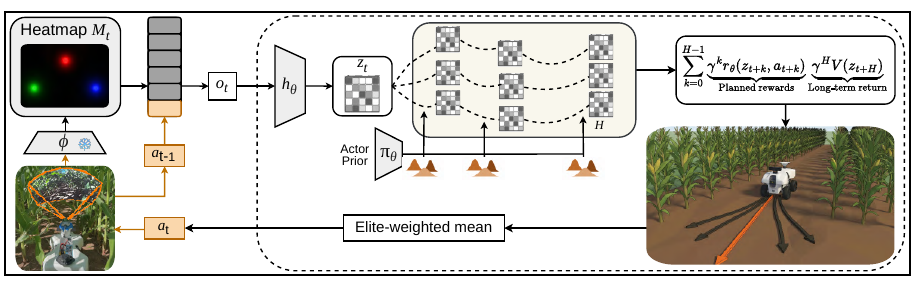}
    \caption{\textbf{LeCropFollow System Overview.} 
(Left) \textbf{Perception:} RGB images are processed through a frozen, self-supervised backbone (RowFollowNet \cite{sivakumar2024cropfollowpp}) to extract semantic heatmaps. These are stacked with the previous action vector to form the encoder input state $o_t$.
(Top Right) \textbf{Training:} The Latent Encoder $h_\theta$, Prior Control Policy $\pi_\theta$, World Model $d_\theta$, Reward $r_\theta$, and Value Function $Q_\theta$ are trained via Reinforcement Learning (following TD-MPC2 \cite{hansen2023tdmpc2}) in a simplified simulation, learning to encode traversability from randomized colored cylinders. 
(Center \& Bottom) \textbf{Inference:} During real-world deployment, the system operates in a zero-shot manner. The pre-trained encoder $h_\theta$ projects observations into the latent space $z$, where MPPI samples candidate trajectories. The optimal action $a$ is selected by maximizing the learned value function over these latent predictions, bridging the sim-to-real gap without online fine-tuning.}
    \label{fig:system_overview}
\end{figure*}

\subsection{Problem Formulation}

We model the under-canopy navigation task not only as a reaction to immediate stimuli but as a local trajectory optimization within a Partially Observable Markov Decision Process (POMDP). The problem is defined by the standard tuple $(\mathcal{S}, \mathcal{A}, \mathcal{T}, \mathcal{R}, \Omega, \mathcal{O}, \gamma)$, where $\mathcal{S}$ is the unobservable state space, $\mathcal{A}$ the action space, $\mathcal{T}(s' \mid s, a)$ the transition dynamics, $\mathcal{R}$ the reward function, $\Omega$ the observation space, $\mathcal{O}(o \mid s)$ the observation model, and $\gamma \in [0,1)$ the discount factor.

The true environment state is not directly accessible; onboard sensors such as the camera provide only a partial projection of the scene, from which task-relevant quantities like the robot's pose relative to the row centerline and the geometric pattern of the plantation cannot be directly recovered. We therefore bypass explicit state estimation and operate on high-dimensional observations $o \in \Omega$, which a learned encoder $h_\theta$ maps to a compact latent representation $z \in \mathcal{Z}$.

A policy $\pi_\theta(z_t)$ is then trained jointly with a world model $d_\theta$ to support planning, following TD-MPC2~\cite{hansen2023tdmpc2}. The objective is to find an optimal action sequence $\mathbf{a}_{t:t+H}$ over a finite planning horizon $H$ that maximizes the expected return in (\ref{eq:planning_objective}), where planned rewards from the world model refine the actions proposed by the policy $\pi_\theta(z_t)$, and a terminal value estimate ensures long-term stability beyond the planning horizon. Only the first action of the optimized sequence is executed, and the optimization is repeated at each control step.

\begin{equation}
a_t = \arg \max_{\mathbf{a}_{t:t+H}} \mathbb{E} \left[ 
    \sum_{k=0}^{H-1} \underbrace{\gamma^k r_{\theta}(z_{t+k}, a_{t+k})}_{\text{Planned rewards}} + 
    \underbrace{\gamma^H V(z_{t+H})}_{\text{Long-term return}}
\right],
\label{eq:planning_objective}
\end{equation}
\noindent
where $r_{\theta}(z,a)$ is the learned reward model and $V(z)$ is the terminal value estimate such that $V(z) \approx Q_\theta(z, \pi_\theta(z))$. The future latent states $z_{t+k}$ are rolled out using the learned dynamics $z_{t+k+1} = d_\theta(z_{t+k}, a_{t+k})$ (Fig.~\ref{fig:system_overview}).

\subsection{State and Action Space}

The raw sensory input is a monocular RGB image $I_t \in \mathbb{R}^{H \times W \times 3}$. To isolate our latent planning hypothesis from any perception advantage, we deliberately match the backbone RowFollowNet of \cite{sivakumar2024cropfollowpp}. $I_t$ is processed through a pre-trained ResNet-18 \cite{he2016deep} into a heatmap tensor $M_t \in \mathbb{R}^{H' \times W' \times 3}$.

\begin{figure}[t]
    \centering
    \includegraphics[width=\linewidth]{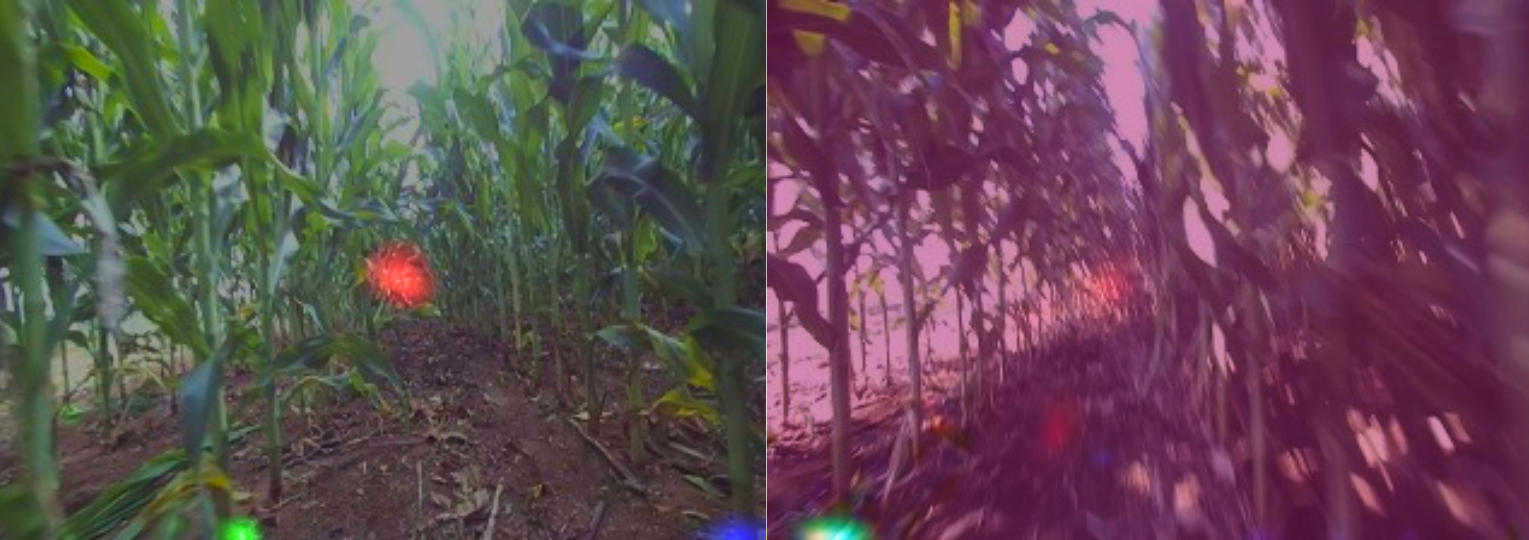}
    \caption{\textbf{Semantic Perception Output.} In-field illustration of the heatmap stacked with the RGB image. (Left) \textbf{High Confidence:} In structured rows, the backbone predicts sharp, compact Gaussian peaks for the Vanishing Point (Red), Left (Green), and Right (Blue) keypoints. (Right) \textbf{High Uncertainty:} In occluded scenarios, predictions become diffuse with high spatial variance. The full heatmap tensor, including such dispersion patterns, is consumed directly by the encoder $h_{\theta}$ without explicit reduction.}
    \label{fig:heatmap}
\end{figure}

Formally, $M_t$ represents the pixel-wise probability density for three semantic classes: the Vanishing Point (red), the Left Row (green), and the Right Row (blue). CropFollow++~\cite{sivakumar2024cropfollowpp} treats the argmax of the spatial distribution as the keypoint for each class. As illustrated in Fig.~\ref{fig:heatmap}, occlusions or irregular rows produce diffuse distributions with high spatial variance. The dispersion of $M_t$ is, by construction, a measure of keypoint-location uncertainty. Rather than extracting either the argmax keypoint or the variance as an explicit scalar parameter, our framework feeds the full heatmap tensor to the encoder $h_{\theta}$, leaving the downstream learned components to adapt to the dispersion as part of the input signal.

Additionally, visual data alone is insufficient to infer the robot’s current velocity or steering state. Therefore, we construct the  observation $o_t$ as the encoder input by concatenating the visual observation $M_t$ with the previous action $a_{t-1} \in \mathbb{R}^{2}$:
\begin{equation}
    o_t = [M_t, a_{t-1}] \in \mathbb{R}^{H' \times W' \times 3 + 2}.
\end{equation}

The encoder $h_\theta$ projects this high-dimensional vector into the compact latent state $z_t = h_\theta(o_t)$.

Finally, on the actions side, the robot operates under unicycle kinematics. The action space $\mathcal{A}$ consists of the target linear velocity $v$ and angular velocity $\omega$:
\begin{equation}
a_t = [v_t, \omega_t].
\end{equation}

The policy outputs are normalized and subsequently scaled to the robot's physical limits: the normalized outputs in $[-1,1]$ are affine-mapped to $v_t \in [0.1, 1.0]$~m/s and $\omega_t \in [-0.9, 0.9]$~rad/s, enforcing a minimum forward speed of $0.1$~m/s and a bounded steering of $\pm 0.9$~rad/s. The commanded action is then smoothed by a first-order exponential filter $\tilde{a}_t = 0.3~a_t + 0.7~\tilde{a}_{t-1}$, which attenuates high-frequency jitter before the command reaches the actuators. The inclusion of $a_{t-1}$ in the state input ensures that the policy accounts for actuator limits and transition dynamics between timesteps.

\subsection{Rewards}
\label{sec:rewards}

To prioritize robot safety over velocity tracking, we adopt a logistic gating formulation proposed in \cite{jaeger2025carl}. This mechanism modulates the task incentive based on system stability, ensuring that high velocities are only rewarded when the platform is stable. We define the total reward $r_t$ as:

\begin{equation}
    r_t = \frac{e^{(r_{\text{task}})}}{1 + e^{(p_{\text{stability}} + p_{\text{collision}})}},
    \label{eq:total_reward}
\end{equation}
\noindent
where $r_{\text{task}}$ is a dense term encouraging forward progress, while $p_{\text{stability}}$ and $p_{\text{collision}}$ are penalty terms. If the robot acts unstably or collides, these terms exponentially nullify the task reward. The specific components are defined as follows:

\begin{enumerate}
    \item \textbf{Task Reward ($r_{\text{task}}$):} A dense term defined as \\ $-\lambda_c \cdot (v_t - v^*)^2$, used to encourage the robot to match a target velocity $v^*$;
    
    \item \textbf{Stability Penalty ($p_{\text{stability}}$):} A dense term defined as \\ $\alpha_c \cdot \omega_t^2 + \beta_c \cdot (|v_t - v_{t-1}| + |\omega_t - \omega_{t-1}|)$. This combines a quadratic penalty on angular velocity ($\alpha_c$) to prevent oscillations with a smoothness penalty ($\beta_c$) that discourages abrupt changes in both linear and angular velocities;
    
    \item \textbf{Collision Penalty ($p_{\text{collision}}$):} A sparse term defined as \\ $\gamma_c \cdot \mathbb{I}(c_t = \text{True})$. If the robot enters a collision state ($c_t$), the reward drops significantly (where $\gamma_c \gg \alpha_c, \beta_c, \lambda_c$).
\end{enumerate}

Hyperparameters $\alpha_c, \lambda_c, \gamma_c \text{ and } \beta_c$ control gating sensitivity and were set following the protocol in \cite{jaeger2025carl}; specifically $\lambda_c = 10$, $\alpha_c = 1.2$, $\beta_c = 0.6$, and $\gamma_c = 30$.

\subsection{Planning with World Models}

We formulate the control problem as trajectory optimization within the learned latent manifold, solved via Model Predictive Path Integral (MPPI) control~\cite{williams2017model}. At each step $t$, given the current observation embedding $z_t$, we sample $N$ perturbed action sequences from a Gaussian distribution centered on rollouts from the learned policy prior $\pi_\theta(z_t)$ to focus the search around promising regions. Each such sequence is then rolled out through the latent dynamics $d_\theta$ over a horizon $H$, and the resulting trajectory is scored against the objective in~\eqref{eq:planning_objective}, where the infinite-horizon return is approximated by appending a terminal value $V(z_{t+H}) \approx Q_\theta(z_{t+H}, \pi_\theta(z_{t+H}))$ to the cumulative reward. The executed action $a_t$ is computed as the reward-weighted average of the top-K elite sequences, applied in a receding horizon scheme that re-plans against the latest observation $o_t$ at every control step.

We assume that local discontinuities such as gaps and severe occlusions are bounded and transient. Under this assumption, the learned world model enables real-time replanning without relying solely on the implicit representations of the policy, which are likely to be biased due to limited exposure to rare events such as gaps. Finite-horizon planning allows the agent to evaluate candidate trajectories beyond such discontinuities instead of reacting only to degraded observations, while the terminal value provides a long-term consistency signal that stabilizes the planned trajectory and preserves coherence.

\subsection{Training Environment}

We implemented the training environment in Gazebo~\cite{koenig2004design} using simple geometric primitives to emulate crop rows. We defined the termination condition as a hard constraint: the episode resets immediately upon any physical collision between the robot and an obstacle, detected directly through Gazebo's physics engine with heuristic contact events. The obstacle environment consists of cylinders spaced at 0.75m matching the row spacing standard of commercial corn plantations and the mechanical dimensioning of the TerraSentia platform \cite{debruin2025breaking}, which are both design assumptions from the deployment scenarios. We deliberately keep the visual representation simple, using untextured cylinders with a randomized color distribution of 80\% green, 10\% red, and 10\% blue.

\begin{figure}[h]
    \centering
    \includegraphics[width=\linewidth]{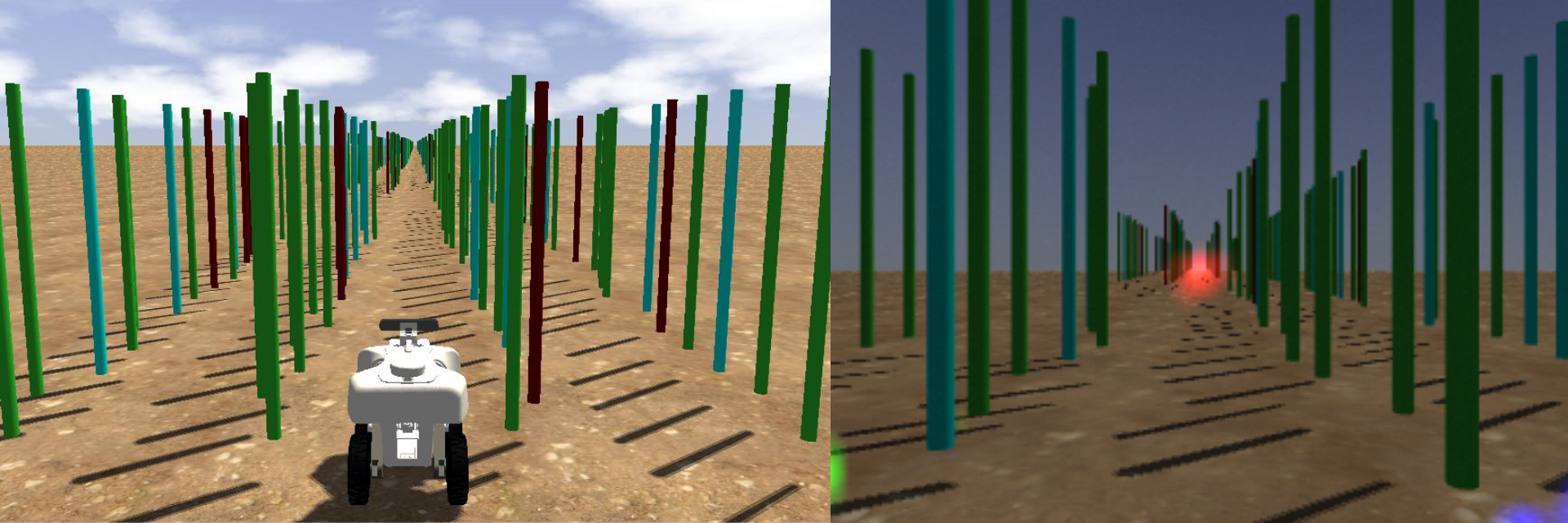}
\caption{\textbf{Simulation Training Environment.} To facilitate rapid RL training, we utilize simplified geometric primitives in Gazebo. (Left) External view showing the robot navigating between cylinder-based rows with 0.75m spacing and randomized color distribution. (Right) The resulting egocentric RGB view overlaid with the predicted heatmap in simulation.}
    \label{fig:simulation}
\end{figure}

Using simple shapes significantly reduces the computational load for rendering and collision physics, accelerating data collection during RL training. More importantly, we treat simulation as a lower-bound out-of-distribution case for the frozen heatmap backbone (Fig.~\ref{fig:simulation}), exposing the downstream encoder $h_\theta$ to noisier and more diffuse activations than those often encountered at deployment. Training under this harder perceptual regime is a key component of our sim-to-real recipe, and empirically, the backbone's heatmap outputs still provide a sufficient learning signal for the policy to converge. The randomized color distribution further reinforces robustness by activating all RGB channels across crop visuals.

Each episode terminates on collision, on reaching a $120$~m forward-distance reset threshold, or at a $500$-step budget, whichever occurs first, so episodes span long continuous rows that expose the policy to extended traversals and to repeated passes through perceptually distinct regions. A short grace window of roughly ten steps at the start of each episode exempts spawn transients from the termination check, preventing spurious resets before the robot settles into the corridor. 

All training hyperparameters and algorithm adaptations are available with source code and training reports.

\begin{figure*}[t]
    \centering
    \includegraphics[width=\linewidth]{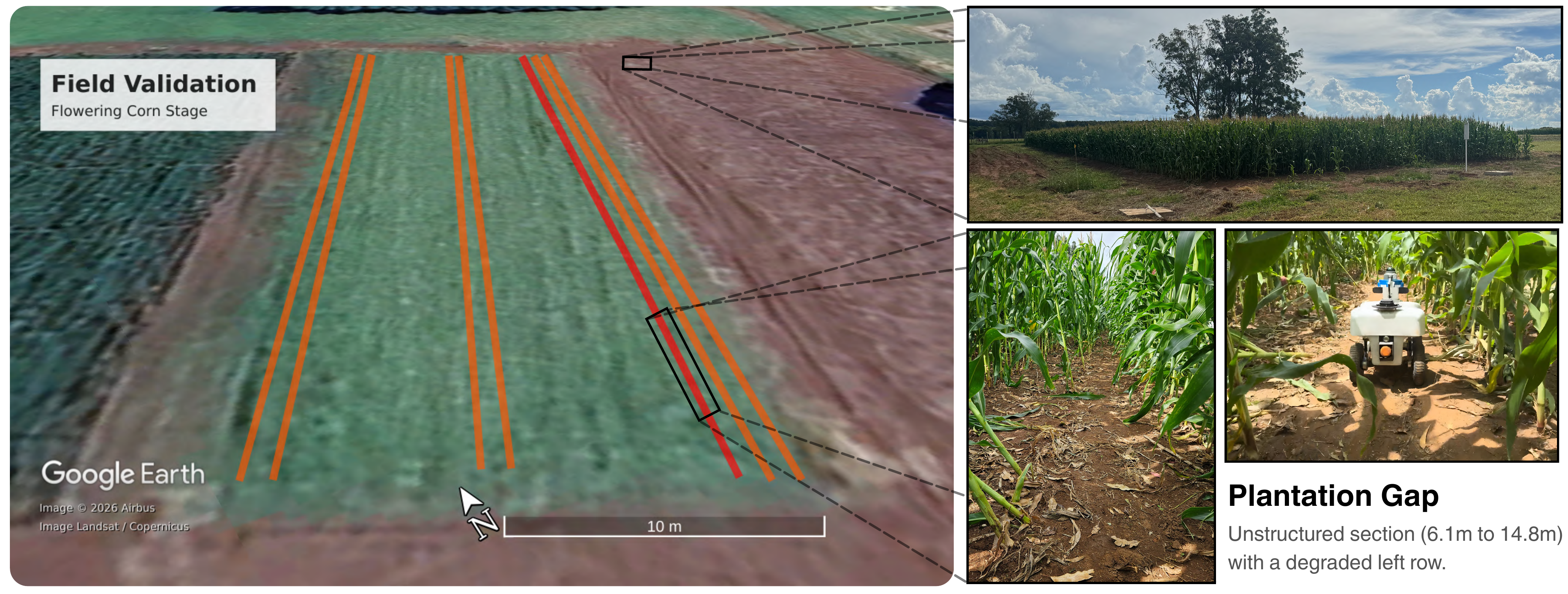}
    \caption{\textbf{Field Validation Environments.} 
    Top-down aerial view of the experimental corn plantation during the Flowering Stage (Source: Google Earth, Airbus, Landsat/Copernicus). 
    The figure highlights the three distinct testing environments evaluated in Tables \ref{tab:lecropfollow_runs} and \ref{tab:corn_metrics}: Left Border, Center, and Right Border rows. Specific focus is drawn to the \textbf{Plantation Gap}, an unstructured section spanning from 6.1~m to 14.8~m (an 8.7~m discontinuity) with a degraded left row. Gap-traversal experiments span both Flowering and Harvested stages within the same plantation morphology, and Harvested rows additionally serve as a generalization probe for LeCropFollow.}
    \label{fig:field-runs}
\end{figure*}

\section{RESULTS}

We evaluated \textbf{LeCropFollow} on two late-stage corn settings. Our experiments specifically targeted unstructured scenarios characterized by irregular planting and frequent occlusions. We provide a categorical analysis of collision modes to rigorously delineate the operational boundaries and failure cases of our method and the baselines.

All field data are open-sourced to ensure reproducibility.

\subsection{Experimental Setup}

\begin{table}[t]
\caption{TD-MPC2 Model and Planning Hyperparameters}
\label{tab:hyperparams}
\centering
\begin{tabularx}{\columnwidth}{XlXl}
\toprule
\textbf{Parameter} & \textbf{Value} & \textbf{Parameter} & \textbf{Value} \\ 
\midrule
MLP Width          & 512  & MPPI Iterations         & 3     \\
MLP Depth          & 3    & Num. Samples ($N$)      & 256   \\
Batch Size               & 512        & Temperature  & 0.50   \\
SimNorm Dim              & 8          & Num. Elites ($K$)       & 48    \\
Num. Q-functions         & 5          & Num. Policy Traj.       & 16    \\
Learning Rate            & $3.75\times10^{-4}$ & MPPI $\sigma_{min}$ & 0.05 \\ 
Buffer Size              & $1\times10^{6}$ & MPPI $\sigma_{max}$ & 2.0   \\
Episode Length           & 500  & Discount ($\gamma$) & 0.99 \\
Seed Steps               & 5500 & Policy Prior Coef.       & 0.20 \\
\bottomrule
\end{tabularx}
\end{table}

Experiments utilized the TerraSentia skid-steer robot (EarthSense Inc.), instrumented with wheel encoders, a 6-DoF IMU, a ZED 2i camera (right monocular stream), and a Livox Mid-360 LiDAR to support the \textbf{CROW}~\cite{affonso2025crow} baseline. We benchmarked against \textbf{CROW} and \textbf{CropFollow++}~\cite{sivakumar2024cropfollowpp} using official public checkpoints, with all systems running at 20~Hz with a target velocity of $v=0.9$~m/s. Notably, CropFollow++ shared our method's frozen heatmap-perception backbone, which emits a $56 \times 80 \times 3$ semantic heatmap over the Vanishing-Point, Left-Row, and Right-Row channels.

Both baselines additionally relied on Direct LiDAR-Inertial Odometry (DLIO)~\cite{chen2022dlio}, running on the Livox Mid-360 and the onboard IMU, to supply the ego-motion estimate their control stacks operate on. In contrast, ours is purely reactive, consuming only the monocular stream and the previous commanded action $a_{t-1}$, and therefore requires no odometry, mapping, or LiDAR at inference.

\textbf{LeCropFollow} is implemented as a lightweight 5M-parameter MLP with Mish activations comprising five learned components ($h_{\theta}, \pi_{\theta}, d_{\theta}, r_{\theta}, Q_{\theta}$), following \cite{hansen2023tdmpc2} report and Table \ref{tab:hyperparams} adaptations. The model was trained for 120k steps in 11.4 hours on an NVIDIA RTX A2000 and deployed on an NVIDIA Jetson Orin Nano, where the policy and planner run asynchronously, throttled to the camera's 20~Hz frame rate.

\subsection{Performance Analysis}

\begin{table}[t]
\centering
\caption{LeCropFollow Field Trials (Flowering Stage -- 12 Runs)}
\label{tab:lecropfollow_runs}
\begin{tabular}{ccc} 
\toprule 
\textbf{Run} & \textbf{Collisions} & \textbf{Max Dist. w/o Col. [m]} \\
\midrule 
1  & 8 & 18.1 \\
2  & 4 & 38.9 \\
3  & 7 & 16.5 \\
4  & 8 & 21.3 \\
5  & 4 & 30.3 \\
6  & 6 & 24.2 \\
7  & 4 & 27.4 \\
8  & 4 & 26.4 \\
9  & 3 & 38.7 \\
10  & 3 & 35.3 \\
11  & 4 & 31.3 \\
12  & 6 & 26.9 \\
\midrule
\textbf{Average} & \textbf{5.1} & \textbf{27.9} \\
\bottomrule
\end{tabular}
\end{table}

\begin{figure*}[t]
    \centering
    \includegraphics[width=\linewidth]{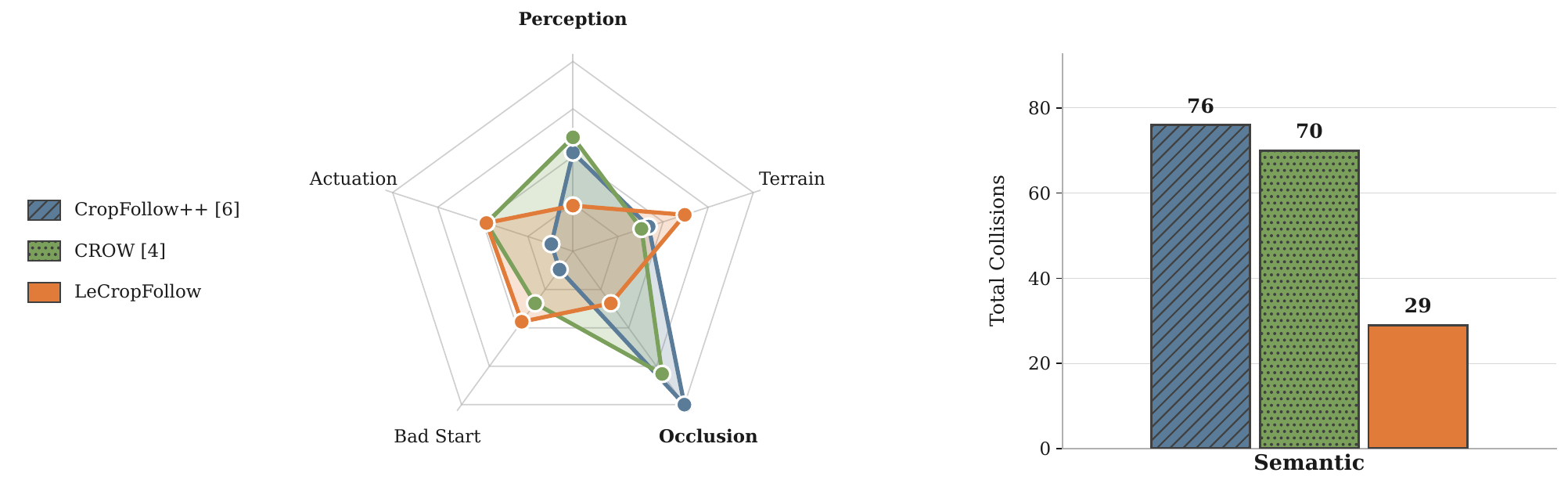}
    \caption{\textbf{Failure Mode Analysis.} 
    (Left) \textbf{Failure Distribution:} Comparative breakdown of failures causes across all methods during field trials. 
    (Right) \textbf{Semantic vs. Physical Trade-off:} Failures are aggregated into Semantic (Perception, Occlusion) and Physical (Actuation, Bad Start, Terrain) categories. \textbf{LeCropFollow reduces Semantic Failures by 2.4$\times$} (29 vs 70) compared to baselines.}
    \label{fig:failure_analysis}
\end{figure*}

\begin{table}[t]
\caption{Comparative Field Trials (Flowering Stage -- 12 Runs Each)}
\label{tab:corn_metrics}
\centering
\begin{tabularx}{\columnwidth}{Xcc} 
\toprule
\textbf{Method} & \textbf{Collisions} & \textbf{\begin{tabular}[c]{@{}c@{}}Max Dist.\\ w/o Col. [m]\end{tabular}} \\
\midrule
CropFollow++ \cite{sivakumar2024cropfollowpp}   & 5.3 $\pm$ 1.6 & 38.9 \\
CROW \cite{affonso2025crow}           & 6.5 $\pm$ 1.4 & 37.9 \\
\addlinespace
\textbf{LeCropFollow} & \textbf{5.1 $\pm$ 1.8} & \textbf{38.9} \\
\bottomrule
\end{tabularx}
\end{table}

We evaluate zero-shot capability across two late-season configurations of the same corn plantation (Fig. \ref{fig:field-runs}). The Flowering Stage spans the transition from late vegetative (VT) to reproductive silking (R1), where dense lower-canopy foliage produces frequent sensor occlusion. The Harvested Stage follows grain collection, exposing residual stalks, bare soil, and a substantially different photometric profile. Both stages share the same row morphology and terrain-induced navigational challenges. Head-to-head comparisons are conducted in the Flowering stage with matched sample sizes per method, while the Harvested Stage serves as a generalization stress probe for LeCropFollow on unstructured scenarios.

Table \ref{tab:lecropfollow_runs} details the overall performance of the system in 12 runs on the Flowering stage only, each traversing a row length of 74.6~m with 0.75~m of internal spacing. To mitigate environmental variability, experiments were conducted in alternating directions and spanned the extremes of the plantation, as illustrated in Fig. \ref{fig:field-runs}. Upon a collision, the robot was manually reset to the row center at the failure point to complete the run.

As reported in Table \ref{tab:corn_metrics}, LeCropFollow performs comparably to established state-of-the-art methods, matching the best baseline on maximum gap-free distance. While both baseline controllers were fine-tuned to fit experimental conditions, ours was deployed in a fully zero-shot manner.

\begin{figure}[t]
    \centering
    \includegraphics[width=\linewidth]{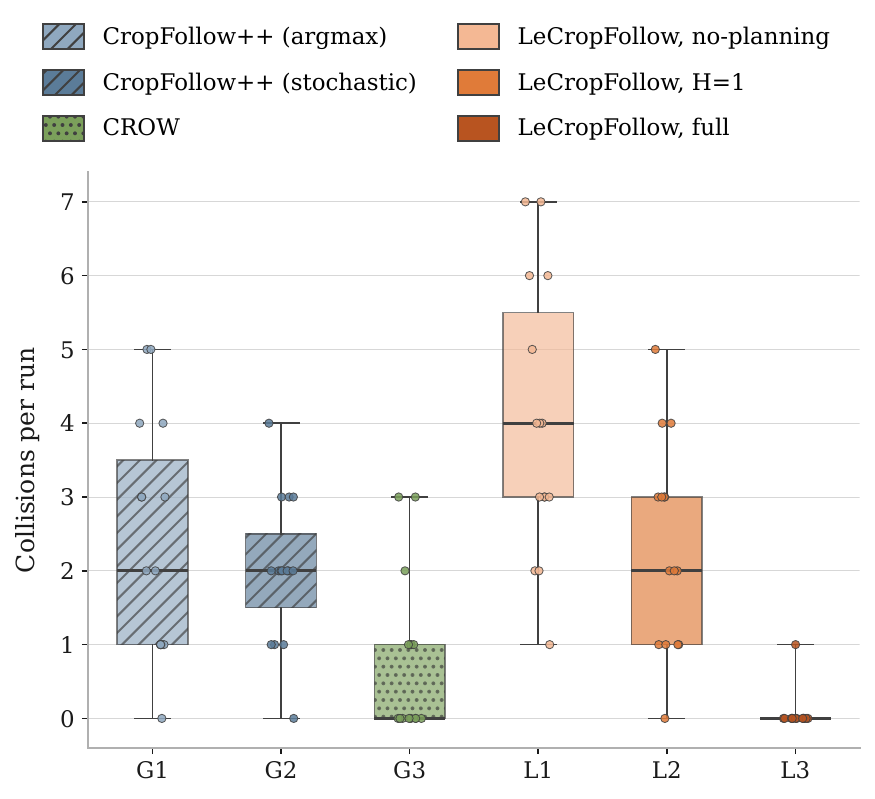}
    \caption{\textbf{Gap Traversal Ablation.} Collisions per run over fifteen traversals of the 8.7~m gap, grouped into Geometric baselines (G1--G3) and Learned variants of our method (L1--L3). The full formulation achieves the lowest median and tightest spread, while removing the planner or shortening the horizon degrades performance below the geometric group.}
    \label{fig:ablation}
\end{figure}

Generally, all methods degrade noticeably relative to their original reports, reflecting the reality gap introduced by our late-season deployment conditions: terrain irregularities and severe canopy occlusions penalize performance regardless of controller optimization, unlike the flatter, more uniform fields typically used in the literature. Under an additional twelve Harvested-stage runs, LeCropFollow yields $4.8 \pm 1.5$ collisions and $28.4 \pm 5.7$~m maximum collision-free distance. A two-sided Mann-Whitney U test finds no significant difference from the Flowering distribution for either metric ($U=63$, $p=0.62$ for collisions, $U=68$, $p=0.74$ for distance), indicating that the policy is not implicitly fitted to a single perceptual configuration of the field. In this adverse context, the parity across all three methods is itself a meaningful outcome, validating LeCropFollow as a competitive alternative to established LiDAR and vision-based approaches.

\subsection{Unstructured Resiliency Analysis}

To isolate the impact of unstructured environments, we selected a row containing an 8.7~m gap on the left side (Fig.~\ref{fig:field-runs}) and conducted fifteen runs per method in alternating directions, pooled across Flowering and Harvested stages since the discontinuity geometry is invariant to crop stage. We present collisions-per-run, where zero collisions are considered a successful run. Baseline hyperparameters were fine-tuned to their best configuration. 

The comparative results in Fig.~\ref{fig:ablation} separate the methods along the geometric-versus-learned axis. CropFollow++ (G1) degrades substantially in this regime (6.7\% success), as its keypoint estimation produces unreliable references when the vanishing point loses consistency, and the geometric controller drives the robot into the adjacent rows, with failures clustered around 8.1~m into the gap. CROW (G3) retains partial robustness (53.3\%) by leveraging a $3 \times 3$~m point cloud window to track adjacent rows, succeeding when at least one row remains within the LiDAR field of view. Operating on the same visual input as CropFollow++, LeCropFollow (L3) traverses the gap in 14 of 15 runs (93.3\%). In high-uncertainty regions, the policy applies small angular corrections rather than tracking the noisy keypoint references, maintaining a smooth heading until the row structure resumes.

Notably, reducing the horizon to a single step (L2) collapses performance close to CropFollow++ (G1), even with Gaussian sampling (G2), showing that one-step rollouts offer limited advantage over deterministic keypoint tracking. Removing the planner and relying on the policy prior $\pi_\theta$ (L1) yields the worst performance overall. The full formulation (L3) outperforms all variants, indicating that planning over the learned world model is the primary source of gain.

\subsection{Failure Mode Analysis}

To understand the operational limits of each method, we aggregated every experimental trial and performed a manual post-hoc analysis with onboard telemetry to attribute the primary cause of collision for every failure case:

\begin{itemize}
    \item \textbf{Perception Error:} Incorrect semantic estimation resulting in minimal corrective control signals ($<0.2$~rad/s) prior to collision, occurring in both continuous rows and discontinuities such as plantation gaps;
    \item \textbf{Occlusion:} Loss of visual tracking due to sensor obstruction by foliage, distinguished from previous by high-uncertainty perception and heading spikes ($>45^\circ$);
    \item \textbf{Actuation:} Inability to track the reference trajectory, quantified by actuator saturation or high-variance control outputs ($>0.6$~rad/s);
    \item \textbf{Bad Start:} Immediate divergence ($<5$~m) caused by suboptimal initial pose;
    \item \textbf{Terrain:} Aggressive heading deviations triggered by ground irregularities or fallen stalks, identified by unusual IMU roll/pitch spikes ($>25^\circ$).
\end{itemize}

To isolate the navigation policy from platform constraints, we classify Perception Error and Occlusion as \textbf{Semantic Failures}. This category represents the core environmental understanding capability addressed in this work (Fig. \ref{fig:failure_analysis}). In contrast, failures resulting from mechanical vehicle-terrain dynamics (Bad Start, Terrain, Actuation) are classified as \textbf{Physical Failures}.

Our primary hypothesis is that geometric methods suffer from information over-compression in unstructured environments; the experimental data strongly supports this. While CropFollow++ yields high-confidence divergence by fitting geometry to noise within plantation gaps, \textbf{LeCropFollow reduces semantic failures by 2.4$\times$} by treating the heatmap's spatial dispersion as part of the input signal.

Despite our method critically excelling in semantically challenging conditions, we report physical failures as the limitation of our current work. Even with a successful zero-shot deployment, the severe uneven terrain stressed the physical platform in non-linear ways, which in many cases induced a heading spike and, ultimately, a collision. The integration of an explicit (or learned) closed-loop controller could improve the mechanical system feedback in those conditions, providing a clearer path for future iterations. Other promising extensions include adding complementary sensing such as depth within the same latent-planning framework.

\section{CONCLUSION}
This work introduces \textbf{LeCropFollow}, a visual navigation framework that circumvents the limitations of explicit geometric modeling in unstructured agricultural fields. By coupling self-supervised semantic heatmaps with a latent-space world model, our framework plans trajectories directly over the uncompressed heatmap signal, preventing the information loss associated with deterministic state estimation. Extensive field experiments on straight rows of corn crops demonstrate that this representational shift enables zero-shot transfer from simplified simulations to real-world deployment, significantly reducing failure rates in plantation gaps compared to geometric baselines. Despite current limitations on physical feedback and gap-geometry scope, we hope our work leads to further research on unstructured field robotics with adaptable learning.

\section*{Acknowledgment}
The authors thank Davide Jarik De Rosa and Jo\~ao H. Al\'essio for their dedicated assistance throughout the field deployments, where their help with robot operation, data collection, and on-site logistics was essential to completing the experimental campaign. The authors also gratefully acknowledge Embrapa Instrumentation (S\~ao Carlos, SP, Brazil) for providing the facilities and field sites where the experiments were conducted, and the Funda\c{c}\~ao de Apoio \`a F\'isica e \`a Qu\'imica (FAFQ) for the operational and administrative support. The authors further acknowledge the Mobile Robotics Group at the Center for Robotics (CRob), EESC USP, for the laboratory infrastructure and the robotic platform.

The Article Processing Charge for the publication of this research was funded by the Coordena\c{c}\~ao de Aperfei\c{c}oamento de Pessoal de N\'ivel Superior (CAPES), Brazil (ROR identifier: 00x0ma614). For open access purposes, the authors have applied a Creative Commons CC BY license to any accepted version of the article.

\IEEEtriggeratref{16}

\bibliography{ref}

\bibliographystyle{IEEEtran}

\vfill

\end{document}